KAUNAS UNIVERSITY OF TECHNOLOGY

Vytautas Perlibakas

**COMPUTERIZED FACE DETECTION AND RECOGNITION**

Summary of doctoral dissertation

Physical sciences, Informatics (09 P)

Kaunas, 2004

KAUNO TECHNOLOGIJOS UNIVERSITETAS

Vytautas Perlibakas

**KOMPIUTERIZUOTAS VEIDO DETEKTAVIMAS IR ATPAŽINIMAS**

Daktaro disertacijos santrauka

Fiziniai mokslai, informatika (09 P)

Kaunas, 2004

# Introduction

The dissertation analyses computerized methods of face detection, analysis and recognition in order to increase accuracy and speed of these methods, and in the future use these developed methods in automatical face recognition systems.

**Relevancy of the doctoral dissertation.** Computerized face detection and recognition is the problem of computer vision, and it is investigated more than ten years. Face recognition is also one of biometric technologies (such as voice, fingerprints, eye iris recognition), which could be used for human identification. Investigations of biometric technologies, including face recognition, are performed in order to create more reliable means for data protection, and develop new technologies for terrorism prevention. Face recognition allows to perform passive identification in complex environment, so it could be used for protection of state frontiers and other territories. Face detection, analysis and recognition methods also could be used in criminalistics, film industry, medicine, for development of intelligent home environment, and in many other fields. Plenty of potential applications stimulate development of these methods. Because face detection and recognition is influenced by many factors (change of face itself and its surrounding environment), these problems are complex and not fully solved. Modern computer technique allows gathering in the databases face images of many persons, so there is need of methods, allowing to perform image-based search in these databases. Also there is need of developing fast detection and recognition methods, corresponding to the speed of today's computers.

**The aim of the work.** The aims of this work are: a) to develop new or modify existing face detection, analysis and recognition methods, in order to increase their speed and accuracy; b) investigate the problems of face features and face contour detection.

**The goals are as follows:** 1. Investigate existing and propose new or modified methods for the detection of face, face features, and face recognition, that would allow to achieve higher speed and accuracy than existing methods. 2. Perform the following investigations: a) investigate influence of the choice of distance measure to the recognition accuracy of the discrete transforms –based face recognition methods; b) investigate the methods for combining face recognition algorithms in order to increase recognition speed and accuracy; c) investigate and modify methods for fast face pre-detection, based on contours, mathematical morphology, and image patterns; d) investigate and modify methods for face features detection, based on mathematical morphology, image projections, and image patterns; e) investigate and modify methods for exact face contour detection based on active contours. 3. As the result of performed investigations, formulate recommendations for future research and development of computerized face detection and recognition methods.

**Used research methods.** In the research are used methods of digital image processing, mathematical statistics, mathematical morphology, variational calculus, theory of sets. Algorithms were realized and experiments performed using MATLAB package (environment) and its programming language.

**Scientific novelty of the dissertation.** 1. Performed investigation of discrete transforms –based face recognition methods and evaluated how the choice of the distance measure influences recognition accuracy. 2. Fast normalized correlation between image patterns –based pre-detection method modified and applied for operation with multiple image patterns. Using the proposed modification computations are performed faster than using unmodified method. 3. Developed active contours –based method for exact face contour detection, allowing to detect face contour in still greyscale images and to reduce influence of scene background to contour detection. 4. Developed face recognition method, based on sequential combination of two methods: at first is used Wavelet Packet Decomposition, and to its result is performed Principal Component Analysis. Using the proposed method, training with large number of images is performed faster than using traditional Principal Component Analysis. And recognition accuracy remains almost the same.

**Practical significance of the work.** The results of this work could be used for creation of computerized face analysis and recognition systems for territory and information protection, law enforcement and criminalistics, video indexing and medical diagnostics.

**Defended proposals:** 1. Method for face pre-detection based on normalized correlation between multiple face patterns. 2. Active contours –based face contour detection method. 3. Cumulative recognition characteristics –based method for combining face recognition algorithms. 4. Face recognition method based on Wavelet Packet Decomposition, and Principal Component Analysis.

**Approbation of the dissertation.** Author has published 3 scientific articles in ISI (Institute of Scientific Information) indexed journals, 3 articles in the journals included in the list certified by the Department of Science and Studies, 5 publications in conference proceedings on the theme of the dissertation. Some parts of the research material and results were published in 3 scientific reports.

**Structure and extent of the dissertation.** The dissertation consists of Introduction, 6 chapters, Conclusions, References (306 entries) and Publications on the theme of the dissertation. The volume of the dissertation is 186 pages and includes 23 tables and 26 figures.

# Content of the dissertation

**Introduction.** Introduction presents relevancy of the theme, the main aim and goals of the work, the novelty and possible practical applications of research, defended proposals, approbation of the dissertation, structure and extent of the dissertation.

**Chapter 1.** ("Face detection and recognition methods and their peculiarity") This chapter presents a literature review on the theme of the dissertation. The chapter consists of six sections ("Face detection", "Face features detection", "Face contour detection", "Face recognition", "Image databases", "Development of commercial face recognition systems").

**Chapter 2.** ("Development and investigation of face detection methods") Before detection of the face or facial features, we must find at least approximate position of the face, that is - perform face detection. In order to perform multiscale face detection using templates (patterns) -based face detector, usually initial image is resized and it is formed a pyramid of resized images. Then in each image of the pyramid is performed face detection and detections from different scales are merged to form detections at original image scale. Because using such method we need to classify many templates and usually template-based classifiers are enough slow, many methods were proposed for speed-up of detection and fast initial detection. In this chapter we describe methods for initial face detection and for verification of initial detections.

<u>The first section.</u> ("Initial face detection") The goal of initial face detection (pre-detection) is to find approximate face positions as fast as possible. Then these positions will be verified and corrected using more accurate and possibly slower methods.

*The first subsection.* ("Contours and ellipses based detection") Because the contour of face is similar to the ellipse, for face pre-detection we detected contours and approximated them with ellipses. Pre-detection steps were as follows: 1. Image lighting correction using histogram equalization and removal of noise using adaptive Wiener filter. 2. Image smoothing using Gaussian filter. 3. Edge detection using Canny edge detector. 4. Linking of edges into contours and splitting of branches during linking, removal of short contours. 5. Splitting of contours at the points with high curvature using k-cosines method. 6. Removal of small circle-like objects (contours), corresponding to small scene objects and face features. 7. Smoothing of contours in order to reduce curvature at the ends of contours after breaking. 8. Analysing of contours endpoints and linking of contours with respect to the distance between endpoints and the orientation of contour's ends. Before linking we test if linked object will not be larger than the size of maximal searched face. 9. Removal of long contours similar to straight lines. In order to speed-up detection we try to remove contours that are not likely faces and reduce the number of contours that will be used for grouping and approximation with ellipse. 10. Approximations of contours with ellipses. We approximate only contours that are possibly faces using size and length constraints. For approximation we use direct least squares fitting method that produces namely ellipses (not parabolas or hyperbolas), proposed by A. Fitzgibbon, M. Pilu and R. B. Fisher in "Direct least square fitting of ellipses" (IEEE Trans. on PAMI, vol. 21, no. 5, 1999: pp. 476-480). 11. Approximation of contour pairs with ellipses. If it is worth to pair contours and approximate them we decide using heuristics that evaluate the size of merged contour, the lengths of separate contours and merged contour, curvature of contours with respect to each other. Heuristics are used in order to speed-up detection. 12. Removal of ellipses that are not likely faces. We evaluate the lengths of ellipse axes, rotation angle of the ellipse, approximation error between the approximated contour and ellipse. We define different errors with respect to the length of the contour. After these steps we get possible locations of faces that should be verified using other methods, analysing image parts inside the detected ellipses.

*The second subsection.* ("Modification of the normalized correlation between templates -based detection method") This subsection describes normalized correlation between multiple templates based face pre-detection method. Linear correlation coefficient (normalized correlation) between templates could be calculated using the following formula:

$$R(X,Y) = \frac{\text{cov}(X,Y)}{\sqrt{DX}\sqrt{DY}} = \frac{M(XY) - MX \cdot MY}{SX \cdot SY}, |R(X,Y)| \leq 1,$$

$$MX = \frac{1}{N_1 \cdot N_2} \sum_{i=1}^{N_1} \sum_{j=1}^{N_2} X(i,j), \quad SX = \sqrt{DX}, \quad DX = M(X - MX)^2 = MX^2 - (MX)^2,$$

here $X$, $Y$ - images of the size ($N_1 \times N_2$), $X$ corresponds to initial image, $Y$ corresponds to changing template, $MX$ - empirical mean, $DX$ - variance, $SX$ - standard deviation, cov - covariance. As it was noted by many researchers, direct calculation of normalized correlation at each point of the image using sliding window is computationally expensive. Computation of normalized correlation between image and template could be speed-up by using Fast Fourier Transform (FFT) (for calculation of $M(XY)$) and running sums (for calculation of $MX$ and $MX^2$) as it was proposed by J. P. Lewis in "Fast Normalized Cross Correlation" (Vision Interface, 1995: pp. 120-123).

Because in many applications it is desirable to perform search with multiple templates, we modified fast normalized cross-correlation method in order to perform search with multiple templates faster than using unmodified method multiple times. Normalized correlation parts $MX$, $SX$, corresponding to unchanging image $X$, we calculate only once and store them. Also we pre-calculate FFT of $X$ for faster calculation of $M(XY)$. Let $[I(N_1, N_2)]$ is initial image, $N_1$, $N_2$ - number of rows and columns, $[T_l(M_1, M_2)]$ - searched template, $l = 1, ..., L$, $L$ - number of templates, $M_1$, $M_2$ - number of template rows and columns (all templates are of the same size). Then the algorithm of proposed modified method for calculation of normalized correlation with multiple templates of the same size, using pseudocode could be written as shown in Fig. 1. After performing loop iterations we get matrices of correlation coefficients $R_l$, $R_l(i,j) \in [-1;1]$, corresponding to correlation coefficients between image $I$ and template $T_l$. For description of the algorithm were used the following notations: $C = \text{fft2p}(A, m_1, n_1, m_2, n_2)$ - Fast Fourier Transform of image $A$;

$m_1$, $n_1$ - dimensions of image $A$; $m_2$, $n_2$ - dimensions used for image padding with zeros; $C = \text{ifft2p}(A, m_1, n_1, m_2, n_2)$ - inverse Fourier transform of image $A$; $m_1$, $n_1$ - dimensions of image $C$; $m_2$, $n_2$ - dimensions for pad removal; $C = \text{conv2}(A, B)$ - convolution of image $A$ with template $B$; $C = \text{lsum2}(A, m, n)$ - local sums of $m$ x $n$ windows in whole image $A$ calculated using running sums; $C = A.*B$, $C(i,j) = A(i,j) \cdot B(i,j)$; $C = A./B$, $C(i,j) = A(i,j)/B(i,j)$; $C = A \pm B$, $C(i,j) = A(i,j) \pm B(i,j)$; $C = \sqrt{A}$, $C(i,j) = \sqrt{A(i,j)}$; $C = A\{+,-,\cdot,/\}b$, $C(i,j) = A(i,j)\{+,-,\cdot,/\}b$; $[C](i,j) \equiv C(i,j)$ - element of $C$ at position $(i,j)$, $\forall i, j$.

---

$m1m2 = M_1 \cdot M_2$

$Imx = \text{lsum2}(I, M_1, M_2)/m1m2$; $Imx2 = \text{lsum2}(I.*I, M_1, M_2)/m1m2$

$Isx = \sqrt{Imx2 - Imx.*Imx}$

$Ifft = \text{fft2p}(I, N_1, N_2, M_1, M_2)$

for $l = 1$ to $L$ do

$\quad Tfft = \text{fft2p}(T_l/m1m2, M_1, M_2, N_1, N_2)$

$\quad Isigm1 = \text{ifft2p}(Ifft.*Tfft, N_1, N_2, M_1, M_2)$ % convolution $\text{conv2}(I, T_l/m1m2)$

$\quad tmx = \dfrac{1}{m1m2}\sum_i \sum_j T_l(i,j)$; $tmx2 = \dfrac{1}{m1m2}\sum_i \sum_j [T_l.*T_l](i,j)$

$\quad tsx = \sqrt{tmx2 - tmx \cdot tmx}$

$\quad Isigm2 = tmx \cdot Imx$

$\quad R_l = (Isigm1 - Isigm2)./(Isx \cdot tsx)$

end for

---

**Fig 1** Modified algorithm of fast normalized correlation with multiple image templates

The speed of the algorithm also depends on some realization details. Because the size of the image and the size of templates remains unchanged, we should allocate memory only once. Calculations with arrays should be performed using SIMD (Single Instruction Multiple Data) instructions. Because FFT is calculated multiple times for images of the same size, its calculation could be speed-up by using transform planning, adopting FFT to the used computer architecture. If detection is performed for the images pyramid and resized images are calculated using FFT, these results of FFT also could be used for calculating correlation coefficients.

After calculation of correlation coefficients, further verification is performed at image positions where $R(i,j) > \tau$ (e.g., $\tau = 0.5$). In order to reduce the number of verifications we leave pre-detections where in 3x3 window the number of $R(i,j) > \tau$ values is larger than some threshold $\tau 2$ (e.g. $\tau 2 = 4$). Then we store coordinates of pre-detections and what templates were used to get these pre-detections. Using this information we crop pre-detected image parts for verifications. For the same pre-detection template could be cropped several image parts corresponding to the same or different face features if pre-detection template is similar to several features. For face pre-detection we used 8 templates of the size 25x11. 6 templates represented eyes with different rotation and distance between eye centres, 1 template of forehead, 1 template of upper lip and nostrils.

The second section. ("Verification of initial detections") For verification of initial face detections usually it is used template-based classifier. Verification is formulated as a classification problem with two classes - faces and nonfaces. For classification we used SNoW (Sparse Network of Winnows) classifier. For gathering nonface templates we used bootstrap method. Face templates were generated from more than 1000 face images using rotation +-3°, resizing 95-105%, mirroring. The size of templates is 20x20, templates were normalized using histogram equalization.

*The first subsection.* ("Features used for verification") In order to reduce the influence of lighting for verification we used not only greyscale features of selected templates, but also horizontal and vertical differences between grey values of neighborhood pixels. Let $[T(M_1, M_2)]$ is a template with $M_1$ rows and $M_2$ columns. Then templates of horizontal and vertical differences are calculated using the following formulas:

$T_h(i1,i2) = (T(i1,i2) - T(i1,i2+1) + 255)/2$, $i1 = 1,...,M_1$, $i2 = 1,...,M_2 - 1$;

$T_v(i1,i2) = (T(i1,i2) - T(i1+1,i2) + 255)/2$, $i1 = 1,...,M_1 - 1$, $i2 = 1,...,M_2$,

here $T(i1,i2) \in [0,...,255]$, $T_h(i1,i2) \in [0,...,255]$, $T_v(i1,i2) \in [0,...,255]$ and values are rounded in order to get integer numbers. Also we used average greyscale values of some 20x20 template rectangular regions, presented in Fig. 2. Average greyscale values of these regions are rounded in order to get integer numbers from the interval [0,255]. All features were binarized and passed to the SNoW classifier.

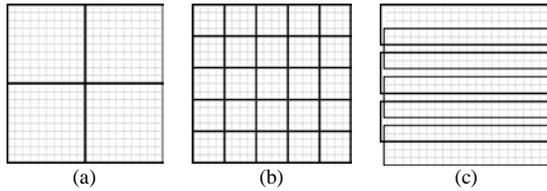

**Fig. 2** Rectangular regions of 20x20 template, for that are calculated mean grey values: a) 4 rectangles (size 10x10 pixels); b) 25 rectangles 4x4; c) 6 horizontal overlapped 20x5 rectangles.

*The second subsection.* ("Merging and elimination of detections") After performing detection in different scales we often get multiple overlapped detection in places where image is similar to the face. Also depending on the detection method we can get slightly shifted overlapped detections of the same scale. These overlappings depend on the used pre-detector and also because classifier is trained using shifted, rotated, rescaled faces. Also we can assume that usually faces in the image do not overlap each other. Using these assumptions we can eliminate some false detections using the following steps: 1. Thresholding (remove alone detections that have smaller number of overlappings than a defined threshold). 2. Merging of overlapped detections (merge overlapped detections of similar sizes into single detection). 3. Removal of left overlappings (remove overlapped detections left after merging).

**Chapter 3.** ("The specifics of face features detection") When face is detected we can perform detection of face features. For this task we used mathematical morphology and templates based methods. In order to speed-up templates based detection we used analysis of horizontal and vertical image projections.

The first section. ("Mathematical morphology based detection of face features"). For the detection of face features (eyes, lips) we used mathematical morphology based method. Detection is performed in following steps:

1. Extraction of dark face regions (like eyes and lips). Extraction of features-like image regions is done using the following morphological bottom hat and thresholding operations:

$$E_1 = Tr((I \bullet S_h) - I), \quad E_2 = Tr((I_{1/2} \bullet S_h) - I_{1/2}),$$
$$E_3 = Tr((I \bullet S_v) - I), \quad E_4 = Tr((I_{1/2} \bullet S_v) - I_{1/2}),$$
$$E = (E_1 + E_2^2 + E_3 + E_4^2)/4, \quad BW = TrM(E),$$

here $I$ – initial image (detected face) after histogram equalization, $I(x,y) \in [0,255]$, $S_h$, $S_v$ - horizontal and vertical structuring elements (sizes 1x7 and 7x1), $I_{1/2}$, $I^2$ - twice smaller and twice larger images than $I$, "$\bullet$" - morphological closing (dilation followed by erosion). Thresholding functions are defined as follows:

$$Tr(I(x,y)) = \begin{cases} I(x,y), I(x,y) \in [v_1, v_2] \\ 0, I(x,y) \notin [v_1, v_2] \end{cases}, \quad TrM(E(x,y)) = \begin{cases} 1, E(x,y) \geq c_1 \overline{E} \\ 0, E(x,y) < c_1 \overline{E} \end{cases},$$

here $v_1=0$, $v_2=255$, $\overline{E}$ - mean grey value of the image $E$, coefficient $c_1$ is determined experimentally. After thresholding we label 8-connected white regions of image BW, where $BW(x,y)=1$.

2. Selection of potential eyes and lips regions using geometrical constraints. Using heuristical rules we decide whether a region could be a potential eye or lips. We resize detected face to a fixed size and define possible minimal and maximal widths of face $min\_face\_w$ and $max\_face\_w$ depending on face detector's accuracy. Then we define $min\_eye\_w = min\_face\_w/10$, $max\_eye\_w = max\_face\_w/2$, $min\_lips\_w = min\_face\_w/5$, $max\_lips\_w = max\_face\_w/2$ and say that region R is a potential eye if region's R parameters satisfy inequalities: $Rw \geq min\_eye\_w$, $Rw \leq max\_eye\_w$, $Rarea \geq 10$, $Rwh\_aspect > 1$, $Rwh\_aspect < 6$, where Rw, Rh is a width and height of the region's bounding rectangle, Rwh_aspect=Rw/Rh, Rarea is a region's area in points. Region R is a potential lips, if region parameters satisfy inequalities: $Rw \geq min\_lips\_w$, $Rw \leq max\_lips\_w$, $Rarea \geq 20$, $Rwh\_aspect > 2$, $Rwh\_aspect < 15$.

3. Generation of eye pairs. We generate potential eye pairs and say that two eyes belong to the same face if they satisfy inequalities: $|R1cy - R2cy| < (R1h + R2h)/2$, $max(R1w,R2w)/min(R1w,R2w) < 1.5$, $max(R1h,R2h)/min(R1h,R2h) < 1.5$, $|R1hist\_mean - R2hist\_mean| < 50$, here Rcy is a region's R centre coordinate y, Rhist_mean - average region's grey value.

4. Generation of features triplets. For each eye pair we try to find lips and form triplets (two eyes and lips) using geometrical constraints and heuristical rules.

5. Verification of features triplets. We remove overlapped triplets using geometrical constraints and verify detected features using templates -based verificator.

Features detected using this method were used for detection of exact face contour.

<u>The second section.</u> ("Templates based eyes detection") For detection of eyes we also used templates-based detector. Faces were detected using normalized correlation based pre-detector and templates-based SNoW (Sparse Network of Winnows) verificator. Then we searched for eyes in the upper part of the detected face. In order to speed-up detection, at first we analysed vertical and horizontal projections (profiles) shown in Fig. 3 and found minimas (intensity walleys). Then at the intersections of minimas in vertical and horizontal projections and some additional points between them we performed multiscale templates-based verification using SNoW classifier. After templates-based verification we get possible locations of eye centre. Then we use accumulator of detections, thresholding, classification of binary objects by area, masking, search of local maximas and found the centre of eye as shown in Fig. 4.

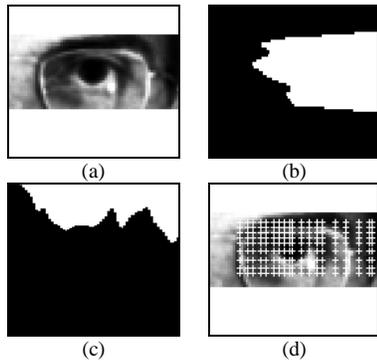

**Fig. 3** Search of potential eye centre: a) initial eye image; b) vertical projection of the eye image; c) horizontal projection; d) intersection points at positions of projection's minimas and additional positions between minimas.

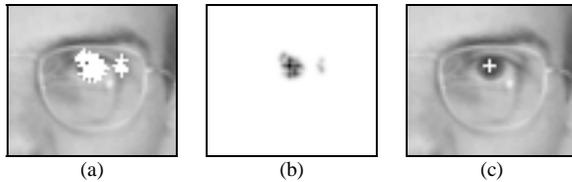

**Fig. 4** Eye centre detection using templates-based verification: a) eye centres after templates-based verification; b) accumulator of detections; c) found eye centre at position of maximal detections number in accumulator array.

**Chapter 4.** ("Development of active contours –based face contour detection method") The face contour is one of the most complex features in a human face image to be extracted. The active contours are probably the best choice for this task. For face contour detection we used variational snakes.

The first section. ("Snake") The active contour model (or snake) is an energy minimizing spline, whose energy depends on the snake's form and position in the image. A snake is found after minimization of the energy functional, which is a sum of internal and external forces with weight coefficients. A snake is defined as a parametric curve $g(s) = (x(s), y(s))$, where x and y are the coordinates of contour points, $s \in [0,1]$. The energy functional, which we wish to minimize, is defined accordingly:

$$E^*_{snake} = \int_0^1 \left( E_{internal}(g(s)) + E_{external}(g(s)) \right) ds .$$

The internal snake's energy: $E_{internal} = (\alpha(s)|g'(s)|^2 + \beta(s)|g''(s)|^2)/2$, where coefficients $\alpha(s)$ and $\beta(s)$ control snake's tension and rigidity (coefficients are determined experimentally). Coefficient values are determined experimentally. We define the external snake's energy as follows: $E_{external} = f(x,y) = EdgeMap(I(x,y))$. How *EdgeMap* is computed we will describe later. If $g(s)$ is a local minima, it satisfies the Euler-Lagrange equation:

$$-\alpha(s)g''(s) + \beta(s)g''''(s) + \gamma \nabla E_{external} = 0 ,$$

here coefficient $\gamma$ is determined experimentally, $\nabla$ - gradient operator.

After discretization in space using the finite differences and appropriate grouping of elements, we can get the following snake's evolution equations:

$$\begin{cases} x^{(t+1)} = (A + \eta I)^{-1}(\eta x^{(t)} - \gamma u(x^{(t)}, y^{(t)})) \\ y^{(t+1)} = (A + \eta I)^{-1}(\eta y^{(t)} - \gamma v(x^{(t)}, y^{(t)})) \end{cases}$$

here $s = 1,...,n$ (n is the number of snake points), $x = (x_1, x_2, ..., x_n)^T$, $y = (y_1, y_2, ..., y_n)^T$, A is an nxn symmetric matrix of coefficients $\alpha$ and $\beta$ (for simplification we use $\alpha(s) = \alpha$ and $\beta(s) = \beta$), $\gamma$ is a weight coefficient of the external force (coefficient value is determined experimentally), $(u(x,y), v(x,y)) = (\partial f(x,y)/\partial x, \partial f(x,y)/\partial y)$ is a potential field, $\eta$ is a time step ($\eta = 1/\Delta t$). A solution is found when evolution equation stabilizes and during time steps the snake's contour remains unchanged. Also we could perform a fixed number of iterations or terminate iterations when after a time step less than a defined percent of the snake's points change their position.

The second section. ("External force field") Traditional snakes do not solve contour detection problem completely and we need to place the initial snake close to the true boundary of the object or else snake will converge to the wrong result. If we place a snake inside the object we must use additional forces expanding the snake. These limitations of traditional snakes could be overcome in snake's equations using not a potential field, but the Generalized Gradient Vector Flow field (GGVF), proposed by C. Xu and J. L. Prince in "Generalized gradient vector flow external forces for active contours" (Signal Processing, vol. 71, no. 2, 1998: pp. 131-139). Using this field, the snake does not need a prior knowledge about whether to shrink or expand toward the boundary and could be initialized far away from the boundary. Generalized gradient vector flow field is defined as a vector field $g(x,y) = (u(x,y), v(x,y))$ that minimizes the following energy functional:

$$E = \iint k(|\nabla f|) |\nabla g|^2 + h(|\nabla f|) |g - \nabla f|^2 \, dxdy ,$$

here $k(|\nabla f|)$, $h(|\nabla f|)$ - weighting functions that control smooth vector field's varying at locations far from the boundaries and conformance to the gradient near the boundaries. If $g(x,y)$ is a minima, it satisfies the Euler-Lagrange equation:

$$-\nabla(k(|\nabla f|) \nabla g) + h(|\nabla f|)(g - \nabla f) = 0 ,$$

$$-\nabla k(|\nabla f|) \nabla g - k(|\nabla f|) \nabla^2 g + h(|\nabla f|)(g - \nabla f) = 0 .$$

For simplification it is used $\nabla k(|\nabla f|) \nabla g = 0$ and a solution is obtained when the following generalized diffusion equation stabilizes:

$$-k(|\nabla f|) |\nabla f|^2 g^{(t)} + h(|\nabla f|)(g^{(t)} - \nabla f) = -\eta \Delta g ,$$

here $\eta = 1/\Delta t = 1$, $\Delta g = g^{(t-1)} - g^{(t)}$. Then the components of the field $g(u,v)$ are as follows:

$$\begin{cases} u^{(t+1)} = u^{(t)} + k(|\nabla f|) \nabla^2 u^{(t)} - h(|\nabla f|)(u^{(t)} - f_x) \\ v^{(t+1)} = v^{(t)} + k(|\nabla f|) \nabla^2 v^{(t)} - h(|\nabla f|)(v^{(t)} - f_y) \end{cases},$$

here $f_x = \partial f(x,y)/\partial x$, $f_y = \partial f(x,y)/\partial y$, initial $u^{(0)} = f_x$, $v^{(0)} = f_y$, weighting functions $k(|\nabla f|) = e^{-(|\nabla f|/K)}$, $h(|\nabla f|) = 1 - k(|\nabla f|)$, coefficient K determines the degree of tradeoff between field's smoothness and conformity to the gradient.

The third section. ("New method for exact face contour detection") This section presents the proposed face contour detection method, for the detection of exact face contour in still greyscale images. Main ideas of the proposed method are as follows:

1. Active contour is initialized inside the face. Contour initialization inside the face allows to reduce the influence of scene's background to the contour detection.

2. Contour initialization is performed using already detected face features. Usage of face features for initialization allows to place initial contour near the true face contour, using our knowledge about faces, proportions of features.

3. As an external force is used generalized gradient field (GGVF). Generalized gradient field allows to expand initial contour without using additional forces and to place it farther from the true face contour, in contrast to when using potential field.

4. Edge map is prepared using detected face features. For calculation of GGVF we must prepare edge map. If we will use all edges of the image, contour will be attracted to facial features, irrespectively if we use GGVF or potential field. So we remove edges, corresponding to internal face part, using detected face features and greyscale information of the internal part of the face. Also we remove edges outside the face.

5. Image filtering is performed with respect to detected face features. Before calculation of edge map, we perform image filtering in order to reduce noise and the number of edges, that do not correspond to the face contour. Because various parts of the face contour have different contrast, we perform filtering with respect to the detected face features.

Face contour is detected in the following steps:

1. Calculation of initial contour. After detection of face features we have corresponding information about right eye, left eye and lips: centres *rec(recx,recy)*, *lec(lecx,lecy)*, *lic(licx,licy)*, widths *rew*, *lew*, *liw* and heights *reh*, *leh*, *lih*. Using this information, we calculate some points around all the features, approximate them with a curve of cubic B-splines and get a contour, that will be used for removal of the face features and as an initial snake. The snake's initialization contour is computed by approximating the following 10 points around the face: *pt3((recx+lecx)/2, recy-dreli\*c3)*; *pt2(recx, recy-dreli\*c2)*, *pt4(lecx, lecy-dreli\*c2)*; *pt1(recx-drele\*c1, recy)*, *pt5(lecx+drele\*c1, lecy)*; *pt10(recx-drele\*c4, licy-dreli/2)*; *pt6(lecx+drele\*c4, licy-dreli/2)*; *pt9(recx-drele\*c5, licy)*, *pt7(lecx+drele\*c5, licy)*; *pt8(licx, licy+dreli\*c6)*, here *(recx, recy)*, *(lecx, lecy )*, *(licx, licy)* - centres of right eye, left eye, lips, *drele=lecx-recx*, *dreli=licy-recy*, *c1=0.5*, *c2=0.2*, *c3=c2+1.0*, *c4=c1\*0.6*, *c5=0.1*, *c6=0.6*. With a curve of cubic B-splines we approximate the sequence of points *pt1*, *pt2*, ..., *pt9*, *pt10*, *pt1* and get a closed contour that is used for snake's initialization and preparation of edge map. Detected face features are also used for image rotation and size normalization.

2. Image lighting normalization and filtering. Initial image *I* is normalize using histogram equalization with 64 bins. Then we filter an upper part of the image above *(licy+2\*lih)* using the median filter with a window of the size 7x7 pixels.

3. Preparation of edge map *EdgeMap*. Initial edge map of the image is calculated using Canny edge detector. For image *I* after lighting normalization and filtering we calculate mean grey value *mgv* of the face without eyes and lips, and create the following binary mask:

$$BWmask(x, y) = \begin{cases} 1, I(x, y) \in [mgv_1, mgv_2] \\ 0, I(x, y) \notin [mgv_1, mgv_2] \end{cases},$$

here $mgv_1=(1-p)*mgv$, $mgv_2=(1+p)*mgv$, *p=0.2*. Also we fill with 1'es parts of *BWmask* using eyes and lips regions. Then we fill holes of the mask and leave only one largest region that masks the face. Then we modify edge map *EdgeMap*. Set to zeros its part below *(licy+6\*lih)*. Remove contours with small lengths (e.g., smaller than the average width of the eyes). Also remove contours inside the central part of the image using mask *BWmask* and contours in the central part of the image using the snake's initialization contour. Remove contours outside the face using ellipse with a centre *((recx+lecx)/2, recy)*, vertical semi-axis of length *(dreli\*2.4)* and horizontal semi-axis of length *(drele\*1.8)* in order to reduce the influence of image background to the face contour detection.

4. Calculation of generalized gradient field and contour detection. For prepared edge map *EdgeMap* we calculate the Generalized Gradient Vector Flow field (GGVF), place the result of these calculations and the initial contour into snake's evolution equation and after fixed number of iterations (determined experimentally) find exact face contour.

Main steps, illustrating face contour detection, are presented in Fig. 5.

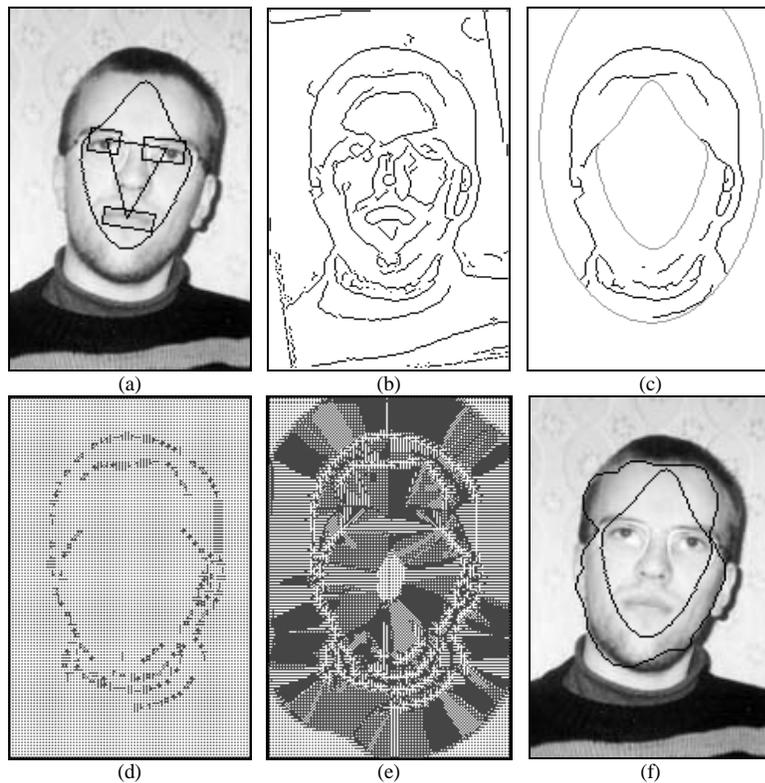

**Fig. 5** Main steps of face contour detection: a) face features and initial snake; b) Canny edge map; c) prepared edge map *EdgeMap* and its preparation contours; d) potential field; e) GGVF field; f) initial snake and detected contour.

**Chapter 5.** ("Face recognition") After face and face features detection we can perform face recognition. In this chapter we describe Karhunen-Loeve, cosine, and wavelet transforms based face recognition methods and their modifications in order to increase recognition speed and accuracy.

<u>The first section.</u> ("Recognition using discrete transforms") Karhunen-Loeve transform (or PCA - Principal Component Analysis) based face recognition method is enough popular and was used by many researchers. Using this method we calculate covariance matrix of the training data and find its eigenvectors and eigenvalues. One of the problems using this method is that calculation of eigenvectors for large matrices is complicated and takes long time. In order to solve this problem we can use incremental eigenspace learning, choose small number of representative face images, split images into small peaces or use other transforms, for example cosine or wavelet transforms. This section presents definitions of these transforms and discusses feature selection and comparison methods.

*The first subsection.* ("Karhunen-Loeve, cosine, and wavelet transforms") This subsection presents main formulas of the discrete transforms, that were used for face recognition:

1. Karhunen-Loeve transform (Principal Component Analysis).

Let $X_j$ be $N$-element one-dimensional image and we have $r$ such images, $j = 1,...,r$. A one-dimensional image [X(N)] from the two-dimensional image [I($N_1,N_2$)] (e.g. face photography) is formed by scanning all the elements of the two-dimensional image row by row and writing them to the column-vector. Then the mean vector, centred data vectors and covariance matrix are calculated:

$$m = \frac{1}{r}\sum_{j=1}^{r} X_j, \quad d_j = X_j - m, \quad C = \frac{1}{r}\sum_{j=1}^{r} d_j d_j^T,$$

here $X = (x_0, x_1,...,x_{N-1})^T$, $m = (m_0, m_1,...,m_{N-1})^T$, $d = (d_0, d_1,...,d_{N-1})^T$, $N = N_1 N_2$. Principal axes are found by calculating eigenvectors $u_k$ and eigenvalues $\lambda_k$ of the covariance matrix C ($Cu_k = \lambda_k u_k$). Because the dimensions $N \times N$ of the matrix C are large even for a small images, and computation of eigenvectors using traditional methods is complicated, dimensions can be reduced using decomposition of the covariance matrix $C = AA^T$, $A = (d_0,...,d_{r-1})/\sqrt{r}$ and calculating eigenvectors of the matrix $A^T A$ (dimensions $r \times r$), if the number of training images $r$ is smaller than the number of image pixels $N$. Found eigenvectors $u = (u_0, u_1,..., u_{N-1})^T$ of the covariance matrix are normed, sorted in decreasing order according to the corresponding eigenvalues, transposed and arranged to form the row-vectors of the transformation matrix $T$. Now any data $X$ can be projected into the eigenspace using the following formula:

$$Y = T(X - m),$$

here $X = (x_0, x_1,...,x_{N-1})^T$, $Y = (y_0, y_1,..., y_{r-1}, 0,...,0)^T$. Also we can perform "whitening" transform $Y = \Lambda^{-1/2} T(X - m)$, $\Lambda^{-1/2} = diag(\sqrt{1/\lambda_0}, \sqrt{1/\lambda_1},...,\sqrt{1/\lambda_{r-1}})$. After whitening, transformed data have zero mean and a covariance matrix given by the identity matrix.

2. Discrete Cosine Transform.

The Discrete Cosine Transform (DCT) is defined as follows:

$$Y(k_1, k_2) = \alpha_1 \alpha_2 \sum_{l_2=0}^{N_2-1} \sum_{l_1=0}^{N_1-1} X(l_1, l_2) \cos\frac{(2l_1+1)\pi k_1}{2N_1} \cos\frac{(2l_2+1)\pi k_2}{2N_2},$$

here $\alpha_1 = \begin{cases} \sqrt{1/N_1}, k_1 = 0 \\ \sqrt{2/N_1}, 1 \le k_1 \le N_1 - 1 \end{cases}$, $\alpha_2 = \begin{cases} \sqrt{1/N_2}, k_2 = 0 \\ \sqrt{2/N_2}, 1 \le k_2 \le N_2 - 1 \end{cases}$, $k_1 = 0,1,..., N_1 - 1$, $k_2 = 0,1,..., N_2 - 1$, and $N_1$, $N_2$ – number of rows and columns of images $X$ and $Y$.

### 3. Discrete Wavelet Transform (DWT).

In this work we used orthogonal compactly supported wavelets. Wavelet transform is defined using basis functions called scaling function $\varphi$ and wavelet function $\psi$: $\varphi(t) = 2^{j/2}\varphi(2^j t - k)$, $\psi(t) = 2^{j/2}\psi(2^j t - k)$, that satisfy relationships

$$\varphi(2^j t) = \sum_k h_{j+1}(k)\varphi(2^{j+1}t - k), \quad \psi(2^j t) = \sum_k g_{j+1}(k)\varphi(2^{j+1}t - k),$$

with corresponding sets of coefficients $\{h_j(k)\}$, $\{g_j(k)\}$, here $j = 0,1,...,J-1$, $k = 0,1,...,K-1$, $t \in R$. Discrete Wavelet Transform (DWT) is calculated using the following iterative equations:

$$a_{j+1}(k) = \sum_{m=0}^{M_j - 1} h(m - 2k)a_j(m), \quad d_{j+1}(k) = \sum_{m=0}^{M_j - 1} g(m - 2k)a_j(m),$$

here $a_0$ - initial image, $a_j$ - approximation at level j, $d_j$ - details at level j, $M_j$ - length of image $a_j$. Low pass filter $h$ and high pass filter $g$ satisfy relationship $g(k) = (-1)^{k+1} h(K - k - 1)$, here $k = 0,1,...,K-1$, $K$ - number of filter coefficients, if $k < 0$ or $k > K-1$, then $g(k) = 0$, $h(k) = 0$. Low pass filter corresponds to scaling function, high pass filter corresponds to wavelet function. In one-dimensional case after performing DWT to the level $J$ we get the result of transform that contains approximation at level $J$ and all details to the level $J$ : $\{d_{j+1}, a_J\}$, $j = 0,1,...,J-1$. In two-dimensional case DWT could be calculated using the following iterative schema:

$$A_j \to LoD_{rows} \to 2\downarrow 1 \to LoD_{cols} \to 1\downarrow 2 \to A_{j+1},$$
$$A_j \to LoD_{rows} \to 2\downarrow 1 \to HiD_{cols} \to 1\downarrow 2 \to D_{j+1,h},$$
$$A_j \to HiD_{rows} \to 2\downarrow 1 \to LoD_{cols} \to 1\downarrow 2 \to D_{j+1,v},$$
$$A_j \to HiD_{rows} \to 2\downarrow 1 \to HiD_{cols} \to 1\downarrow 2 \to D_{j+1,d},$$

here $LoD_{rows}$, $LoD_{cols}$, $HiD_{rows}$, $HiD_{cols}$ - convolutions of rows and columns of the two dimensional image $A_j$ with decomposition filters $LoD$ (low pass) and $HiD$ (high pass), $2\downarrow 1$ - dyadic downsampling of columns and leaving odd indexed columns (indexing starts from 0), $1\downarrow 2$ - dyadic downsampling of rows and leaving odd indexed rows, $A_0$ - initial two dimensional image, $A_j$ - image approximation at level j, $D_{j,h}$, $D_{j,v}$, $D_{j,d}$ - horizontal, vertical, diagonal details at level j, $j = 0,1,...,J-1$, $J$ - number of decomposition levels. After decomposition to level $J$ we get transformation result that contains approximation at level $J$ and all details to the level $J$ : $\{D_{j+1,h}, D_{j+1,v}, D_{j+1,d}, A_J\}$, $j = 0,1,...,J-1$. These results could be written to two-dimensional image $Y$ of the same size as initial image $A_0$.

*The second subsection.* ("Compression of feature vector and recognition") For compression of feature vectors and selection of main features we used variances based criterion. That is we leave desired number of features in vectors $Y$ with highest class variances in spectral domain. In PCA case we leave features corresponding to the largest eigenvalues. In cosine and wavelet transforms case we calculate class variances for feature vectors $Y$ of training images. For recognition we used nearest neighbour classifier. Between feature vector $Y_n$ corresponding to unknown person's face and all feature vectors $Y_i$ of known faces we calculate selected distance $\varepsilon_i = d(Y_n, Y_i)$ and find indice of a known face $s = \arg\min_i[\varepsilon_i]$, $i = 1,...,Q$, here $Q$ is the number of known faces. Then we say that face with feature vector $Y_n$ belongs to a person with indice $s$ and feature vector $Y_s$. For rejection of unknown faces we use threshold $\tau$ and say that face with feature vector $Y_n$ belongs to unknown person if $\varepsilon_s \geq \tau$.

*The third subsection.* ("Distance measures") This subsection presents distance measures that could be used for comparison of feature vectors. Let $X$ and $Y$ be feature vectors of length $n$. Then we can calculate the following distance measures between these feature vectors:

1) Minkowski distance ($L_p$)

$$d(X,Y) = L_p(X,Y) = \left(\sum_{i=1}^{n}|x_i - y_i|^p\right)^{1/p}, p > 0;$$

2) Manhattan distance ($L_1$, city block distance)

$$d(X,Y) = L_{p=1}(X,Y) = \sum_{i=1}^{n}|x_i - y_i|;$$

3) Euclidean distance ($L_2$)

$$d(X,Y) = L_{p=2}(X,Y) = \|X - Y\| = \sqrt{\sum_{i=1}^{n}(x_i - y_i)^2};$$

4) Angle based distance
$$d(X,Y) = -\cos(X,Y),$$

here $\cos(X,Y) = \cos(\alpha) = \dfrac{\sum_{i=1}^{n} x_i y_i}{\sqrt{\sum_{i=1}^{n} x_i^2 \sum_{i=1}^{n} y_i^2}}$, $\cos(X,Y) \in [-1,1]$;

5) Correlation coefficient based distance
$$d(X,Y) = -r(X,Y),$$

here $r(X,Y) = r = \dfrac{n\sum\limits_{i=1}^{n} x_i y_i - \sum\limits_{i=1}^{n} x_i \sum\limits_{i=1}^{n} y_i}{\sqrt{\left(n\sum\limits_{i=1}^{n} x_i^2 - \left(\sum\limits_{i=1}^{n} x_i\right)^2\right)\left(n\sum\limits_{i=1}^{n} y_i^2 - \left(\sum\limits_{i=1}^{n} y_i\right)^2\right)}}$, $r(X,Y) \in [-1,1]$;

6) Weighted angle based distance

$$d(X,Y) = \dfrac{-\sum\limits_{i=1}^{n} z_i x_i y_i}{\sqrt{\sum\limits_{i=1}^{n} x_i^2 \sum\limits_{i=1}^{n} y_i^2}}$$, here $z_i = \sqrt{1/\lambda_i}$, $\lambda_i$ - corresponding eigenvalues;

7) Simplified Mahalanobis distance (numerator of weighted angle-based distance)

$$d(X,Y) = -\sum_{i=1}^{n} z_i x_i y_i .$$

Before calculating distances between PCA feature vectors, we can perform whitening and calculate distances between whitened feature vectors. In the case of other transforms, before calculating distances we can perform centring and standardization of feature vectors.

The second section. ("Development of face recognition method by combining Principal Component Analysis and Wavelet Packet Decomposition") Because usage of PCA -based recognition with large number of training images is complicated, we decided to decompose image into smaller parts using Wavelet Packet Decomposition (WPD) and then perform PCA for decomposed smaller images.

*The first subsection.* ("Wavelet Packet Decomposition") Using the classical wavelet decomposition, the image is decomposed into the approximation and details images, the approximation is then decomposed itself into a second level of approximation and details and so on. Wavelet Packet Decomposition (WPD) is a generalization of the classical wavelet decomposition and using WPD we decompose both approximations and details into a further level of approximations and details. Using WPD we decompose the two-dimensional initial image $A_0^0$ (level $l = 0$) into approximation $A_0^1$, horizontal details $D_{0,h}^1$, vertical details $D_{0,v}^1$ and diagonal details $D_{0,d}^1$ at level $l = 1$. In order to get decomposition at level $l$ we decompose approximations $A_i^{l-1}$ and details $D_{i,h}^{l-1}$, $D_{i,v}^{l-1}$, $D_{i,d}^{l-1}$ into the following approximations and details:

$A_i^{l-1} \to \{A_{4i}^l; D_{4i,h}^l; D_{4i,v}^l; D_{4i,d}^l\}, l > 0$,

$D_{i,h}^{l-1} \to \{A_{4i+1}^l; D_{4i+1,h}^l; D_{4i+1,v}^l; D_{4i+1,d}^l\}, l > 1$,

$D_{i,v}^{l-1} \to \{A_{4i+2}^l; D_{4i+2,h}^l; D_{4i+2,v}^l; D_{4i+2,d}^l\}, l > 1$,

$$D_{i,d}^{l-1} \to \{A_{4i+3}^l; D_{4i+3,h}^l; D_{4i+3,v}^l; D_{4i+3,d}^l\}, l > 1,$$

here $i = 0, ..., (4^{(l-1)} - 1)$. Then at level $l$ we have approximations and details $\{A_i^l; D_{i,h}^l; D_{i,v}^l; D_{i,d}^l\}$. Quaternary tree of such decomposition is shown in Fig. 6 (for $l = 2$). Approximations and details are calculated using low-pass and high-pass decomposition filters and dyadic downsampling using the following schema:

$$A_0^0 \to LoD_{rows} \to 2\downarrow 1 \to LoD_{cols} \to 1\downarrow 2 \to A_0^1,$$
$$A_0^0 \to LoD_{rows} \to 2\downarrow 1 \to HiD_{cols} \to 1\downarrow 2 \to D_{0,h}^1,$$
$$A_0^0 \to HiD_{rows} \to 2\downarrow 1 \to LoD_{cols} \to 1\downarrow 2 \to D_{0,v}^1,$$
$$A_0^0 \to HiD_{rows} \to 2\downarrow 1 \to HiD_{cols} \to 1\downarrow 2 \to D_{0,d}^1,$$

here $A_0^0$ - two-dimensional input image, $LoD_{rows}$, $LoD_{cols}$, $HiD_{rows}$, $HiD_{cols}$ - convolutions of rows and columns of the input two-dimensional image with low-pass and high-pass decomposition filters, $2\downarrow 1$ - dyadic downsampling of columns and keeping odd indexed columns (indexing starts from 0), $1\downarrow 2$ - dyadic downsampling of rows and keeping odd indexed rows, $A_0^1$ - approximation at first level, $D_{0,h}^1$, $D_{0,v}^1$, $D_{0,d}^1$ - horizontal, vertical and diagonal details at first level. In order to get decomposition at level $l$ we use the same steps and decompose approximations $A_i^{l-1}$ and details $D_{i,h}^{l-1}$, $D_{i,v}^{l-1}$, $D_{i,d}^{l-1}$. In Fig. 7 we show an example of the WPD for two-dimensional image at levels $l = 0$ (initial image), $l = 1$, $l = 2$ using Haar wavelets.

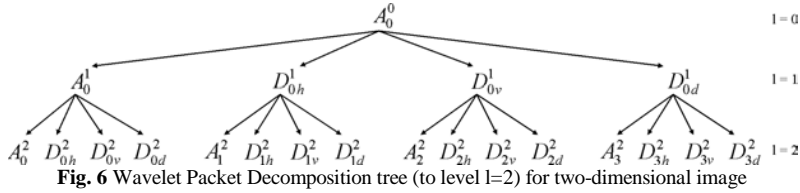

**Fig. 6** Wavelet Packet Decomposition tree (to level l=2) for two-dimensional image

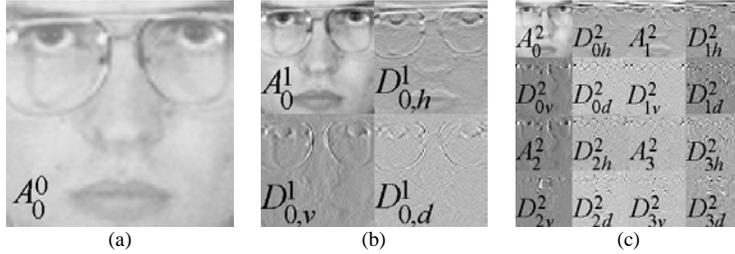

(a)        (b)        (c)

**Fig. 7** Images of approximation (A) and details (D) of the Wavelet Packet Decomposition: a) initial image (approximation at level $l = 0$); b) level $l = 1$; c) level $l = 2$.

*The second subsection.* ("Combining Wavelet Packet Decomposition and Principal Component Analysis") In this subsection we present the proposed face recognition method, based on combination of Wavelet Packet Decomposition and Principal Component Analysis.

Using for recognition traditional Principal Component Analysis -based method, we need to calculate covariance matrix of the training data and then eigenvectors and eigenvalues of this matrix. Because this matrix is large even for small images ($N \times N$, here $N$ is the number of image pixels), training takes long time. If the number of training images is small, we can use the decomposition of the covariance matrix and calculate eigenvectors for matrix $r \times r$, where $r$ is the number of training images. Time complexity of the eigenvectors calculation is $O(n^3)$, $n = \min(N, r)$, here $N$ - number of image pixels, $r$ - number of training images. When the number of images becomes large, we have the same problem eigenvectors calculation for large covariance matrix.

We propose to decompose initial image into $k$ parts using the Wavelet Packet Decomposition and then perform PCA ($k$ times) for these smaller training images (approximations and details). Using the proposed method the time-complexity of the eigenvectors calculation problem is $k \cdot O((N/k)^3) = O(N^3)/k^2$ and is independent from the number of training images, here $k$ is the number of approximations and details. Using the proposed method, training images are added and removed by decomposing them, updating covariance matrices and mean vectors, and recalculating eigenvectors using conventional methods. After the training we get $k$ eigenspaces, and in order to perform recognition we must decompose images into $k$ parts, project them into $k$ eigenspaces and get $k$ feature vectors. Feature vectors are then compressed and compared using a selected distance measure. With what number of training images there is point to use WPD+PCA instead of PCA ? Experiments showed, that if the size of template is 64x64 and $l = 2$, then there is point to use WPD+PCA instead of PCA if the number of training images is larger than 650. If template size is 128x128 and $l = 3$, then there is point to use WPD+PCA instead of PCA if the number of training images is larger than 900. In Fig. 8 is presented training speed-up of WPD+PCA with respect to PCA when template size is 128x128 and $l = 3$.

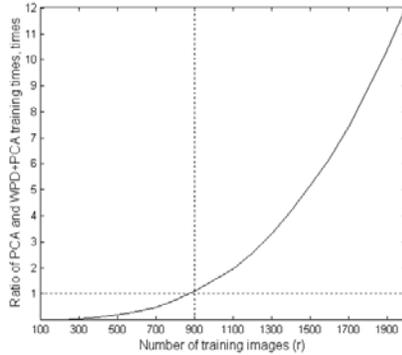

**Fig. 8** Ratio of PCA and WPD+PCA training times (speed-up) with respect to the number of training images ($r$) when using 128x128 images and decomposition into 3 levels

*The third subsection.* ("Compression of composite feature vector and recognition") This subsection discusses the questions of selecting wavelet bases, feature selection and comparison methods for WPD+PCA face recognition method. We choose wavelets experimentally after performing recognition experiments with different orthogonal wavelets. Because after performing WPD+PCA we have $k$ eigenspaces, the question of selecting desired number of features is rather difficult. As in PCA case we should select features corresponding to the largest eigenvalues, but we must decide what number of features in what eigenspace to choose and how to compare selected features: calculate $k$ distances and then combine them or calculate single distance for merged feature vector. After numerous experiments with different feature selection and comparison methods we decided to merge feature vectors into one vector, select features corresponding to the largest eigenvalues of the merged and sorted eigenvalues vector and calculate the selected distance measure between these merged feature vectors. As in PCA case, for recognition we used nearest neighbour classifier and tested different distance measures.

The third section. ("Cumulative recognition characteristics -based combining of face recognition algorithms") One of possible methods to increase face recognition accuracy and speed is to combine several recognition methods. Combining could be done by addition of distance measures with weights, voting of classifiers or combining of results using logical operators, that is parallel combining. We wished to combine algorithms in order to increase recognition speed without loss of accuracy. In order to achieve this goal by combining recognition algorithms, at first recognition should be done using fast recognition method or distance measure and most dissimilar faces rejected in initial stage. Left faces must be compared and recognized using more accurate (although slower) methods (or distance measures). Such method is called a serial combining.

This section presents a proposed method for serial combining of two recognition algorithms. Proposed method allows to calculate cumulative recognition (match) and receiver operating characteristics of the combined method using the distance arrays of the two methods that are combined. Also we describe how to determine (using cumulative characteristics) if there is point to combine two recognition algorithms without performing real experiments with combined method. In order to determine what percent of faces should be selected using first method and then compared using second method we should use cumulative recognition characteristic (CMC) of the first method. This characteristic is determined after testing recognition method with the database of known faces. When choosing methods for combination it is desirable that 100% cumulative recognition rank of the first method be as small as small as possible. Percent (rank) of faces that will be selected by the first method and then compared using the second method is chosen such that this percent be larger (or not smaller) than 100% cumulative recognition rank of the first method. So we can partially guarantee that first one recognition accuracy of combined methods will be not smaller than first one recognition accuracy of the second method. If we wish to combine recognition methods in order to increase recognition accuracy, initial search should be done using a method with small rank values of 100% cumulative recognition. Then left most similar faces should be compared using a method with high first one recognition accuracy. Idea of combining two methods is presented in Fig. 9. Whether there is point to combine two methods we decide accordingly to the goals of combining and the values of the speed and 100% cumulative recognition rank of these methods. In order to calculate cumulative recognition characteristic of combined method, we need to have distance arrays (after performing recognition experiments) of two combined methods, merge these arrays and then calculate cumulative recognition characteristic using this merged array of distances. The speed of the combined recognition method is determined using the speeds of two combined methods.

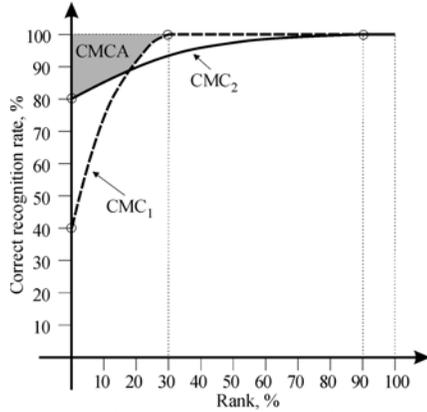

**Fig. 9** Combining of two recognition algorithms using their cumulative match characteristics (CMC$_1$ and CMC$_2$). At first recognition is performed using the algorithm with characteristic CMC$_1$, then some part of images are recognized using the algorithm with characteristic CMC$_2$.

**Chapter 6.** ("Comparative analysis of the experimental results") This chapter presents the experiments and results of face detection, face contour detection, face recognition. Also we compare our results with the results of other researchers and describe face detection and recognition methods used by other researchers.

The first section. ("Face detection experiments and results") Normalized correlation between templates based face pre-detector and SNoW-based verificator was tested using 2391 face images (1196 fa, 1195 fb) from the FERET database created in 1993-1996 years during FERET (Face Recognition Technology) program sponsored by the United States Department of Defence (DoD). For pre-detection we used 8 templates of the size 25x11, for face verification and eyes detection we used templates of the size 20x20. Detection error was measured using the following error measure: $t = 100\% \cdot \max(d_l, d_r)/d_{lr}$, here $d_l$ - Euclidean distance between manually selected and automatically detected left eye centres, $d_r$ - distance between manually selected and automatically detected right eye centres, $d_{lr}$ - distance between the centres of manually selected left and right eye centres. We say that the face was detected correctly if $t < t_{tr}$. If $t_{tr} = 25\%$, achieved detection accuracy is 95.40% with 32 false detections (acceptances). Fig. 10 presents the dependence between desired parameter $t_{tr}$ and the percent of correct detections for $t \leq t_{tr}$.

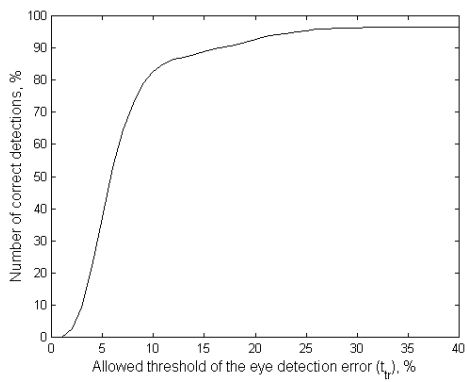

**Fig. 10** Dependence of the number of correct detections (%) from the chosen eye detection error threshold $t_{tr}$ (%)

Also we compared the speeds of pre-detection methods based on fast normalized correlation and modified fast normalized correlation. If the size of images is from 100x100 pixels to 500x500 and the size of templates is from 5x5 to 55x55, modified fast normalized correlation with 5-20 templates is performed 1.2-2.8 times faster than unmodified fast normalized correlation.

The second section. ("Face contour detection results") This section presents the results of face contour detection using snakes. Face contours were detected for 423 face images with correctly detected face features (both eyes and lips). Initial face detection was performed using contours and ellipses based method, face features were detected using mathematical morphology-based method. Images were rotated and rescaled in order to make line connecting eyes horizontal and the distance between eyes equal to 40 points. GGVF field was calculated using *K=0.004* and 60 iterations. Snake was calculated using $\alpha=0.35$, $\beta=0.35$, $\gamma=0.30$, ($\alpha+\beta+\gamma=1$), time step $\eta=1$ and the number of iterations is equal to 250. Contour detection errors were evaluated using the following error measures between manually selected and automatically detected contours:

$$err_1 = 100\% \cdot |(A \cup B) \setminus (A \cap B)|/(|A|+|B|), \quad err_2 = err_1 \cdot (|A|+|B|)/|A|,$$

$$err_3 = (err_{31} + err_{32})/2, \quad err_{31} = D(a,b), \quad err_{32} = D(b,a),$$

$$D(a,b) = \frac{1}{N_a} \sum_{i=1}^{N_a} \min_{j=1,\ldots,N_b} d(a(i), b(i)),$$

here *A*, *B* - sets of points inside face region, contour of region *A* is selected manually, contour of region *B* is detected automatically, $D(a,b)$ is a distance from manually selected contour *a* to contour *b*, $d(\cdot)$ is the Euclidean distance between two points, and $N_a$, $N_b$ is the number of points in contours *a* and *b*. Average contour detection errors are presented in Table 1. As we can see from the table, average distance between manually selected and automatically detected contours is 2.9 points (the distance between eye centres is 40 pt).

Table 1 Average contour detection errors, measured using different error measures

| a, A | b, B | $err_1$, % | $err_2$, % | $err_3$, pt |
|---|---|---|---|---|
| original | final | 6.8 | 14.1 | 2.9 |
| original | initial | 24.8 | 39.8 | 8.8 |

Notation in the table: "original" - manually selected contour, "final" - detected contour using snakes, "initial" - snake's initialization contour.

The third section. ("Face recognition experiments and results") This section describes performed face recognition experiments and presents the results. For comparison of face recognition methods we used the following characteristics of the biometric systems presented in Fig. 11-12: First1 – first one recognition (larger is better), EER – Equal Error Rate (smaller is better), Cum100 – rank (percent) of images needed to extract in order to achieve 100% recognition rate (smaller is better), ROCA – area below ROC (smaller is better), CMCA – area above CMC (smaller is better).

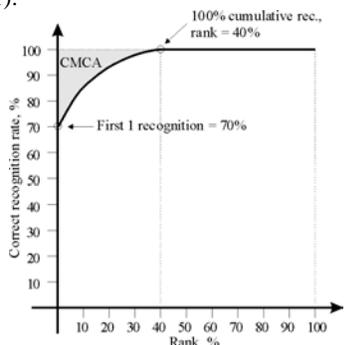 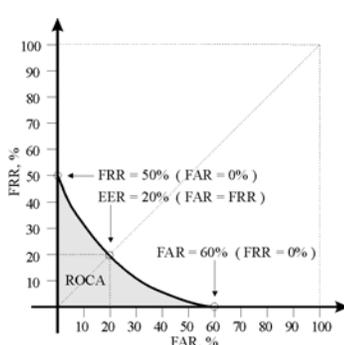

Fig. 11 Cumulative Match Characteristic    Fig. 12 Receiver Operating Characteristic

*The first subsection.* ("Face recognition results using manually detected faces") This subsection describes face recognition experiments using manually detected faces and face features. Because in most freely available face databases the number of persons is small, for recognition experiments we collected photographies of 423 persons (2 images per person - 1 for learning and 1 for testing) from 9 databases. The centres of eyes and lips were selected manually and using these centres we normalized rotation and scale of images, cropped 64x64 templates, performed histogram equalization. At first we performed experiments with PCA in order to find out what distance measures should be used in order to achieve the best recognition results. Experiments with more than 14 distance measures and their modifications using different number of features showed that the best results with respect to different characteristics are achieved using angle between whitened vectors (whitened angle), whitened correlation coefficient, weighted angle. Some results are presented in Table 2.

**Table 2** Recognition using PCA and different distance measures (number of features is 127)

| Method | Cum100, % | CMCA [0..10$^4$] | First1, % | EER, % | ROCA [0..10$^4$] |
|---|---|---|---|---|---|
| Euclidean | 86.1 | 215.20 | 83.22 | 7.33 | 261.63 |
| L1 (Manhattan) | 78.7 | 155.20 | 83.92 | 8.51 | 318.38 |
| Simpl. Mahalanobis | 23.9 | 54.57 | 87.94 | 3.07 | 45.19 |
| Weighted angle | 26.0 | 57.15 | 88.42 | 3.31 | 44.08 |
| Whitened angle | 49.9 | 98.92 | 88.18 | 3.78 | 84.84 |
| Whitened corr. coef. | 52.0 | 101.52 | 88.89 | 3.78 | 87.78 |

Also we tested if the differences between distance measures with respect to Cum100, CMCA, First1, EER, ROCA are statistically significant. We used bootstrap resampling with replacement, ANOVA (ANalysis Of VAriance) and Tukey's HSD (Honest Significant Difference) post hoc test. We used 30% of features (127), the number of bootstrap samples N=2000, significance level $\alpha=0.001$. These tests showed that the mean differences are not statistically significant between simplified Mahalanobis and weighted angle with respect to ROCA, whitened angle and whitened correlation with respect to EER and ROCA, simplified Mahalanobis and whitened angle with respect to First1, weighted angle and whitened angle with respect to First1 and some other distance measures. Many other differences are statistically significant. Significance tests were performed using SPSS 12.0 statistical package.

Experiments using DCT and DWT with different wavelet bases (Daubechies, Symmlets, Coiflets) showed that using Euclidean distance the results are similar to that achieved by PCA and Euclidean distance, but are worse than achieved using PCA and weighted angle based distance. Also the experiments showed that using DCT and DWT better results are achieved using angle between centred, standardized feature vectors or using weighted angle than using Euclidean distance between feature vectors. Some results are presented in Table 3. These experiments showed that recognition accuracy depends on the chosen distance measure.

**Table 3** Recognition accuracy using PCA, DCT, DWT (Daubechies with filter length 12) with 127 features and different distance measures

| Method | Distance measure | Cum100, % | CMCA [0..10$^4$] | First1, % | EER, % | ROCA [0..10$^4$] |
|---|---|---|---|---|---|---|
| PCA | Weighted angle | 26.0 | 57.15 | 88.42 | 3.31 | 44.08 |
| PCA | Euclidean | 86.1 | 215.20 | 83.22 | 7.33 | 261.63 |
| DCT | Euclidean | 90.8 | 259.85 | 82.74 | 6.86 | 301.94 |
| DWT | Euclidean | 92.2 | 272.18 | 82.98 | 7.09 | 307.59 |
| DCT | Angle | 91.7 | 249.82 | 82.51 | 7.33 | 290.94 |
| DWT | Angle | 92.0 | 268.18 | 82.74 | 7.09 | 304.34 |
| DCT | Angle (centred) | 49.4 | 111.25 | 84.16 | 6.15 | 133.14 |
| DWT | Angle (centred) | 49.6 | 116.58 | 83.45 | 6.15 | 138.78 |
| DCT | Angle between standardized | 20.8 | 78.10 | 82.74 | 4.96 | 80.98 |
| DWT | Angle between standardized | 28.1 | 89.00 | 83.69 | 5.20 | 94.78 |

Also we performed experiments using combination of distance measures. Some results are presented in Table 4. In order to increase recognition performance of the weighted angle based distance, at first we performed recognition using simplified Mahalanobis distance and then resorted 25% of most similar images using weighted angle based distance. As we can see from the results, combined method achieved the same First1 recognition as weighted angle based distance and other recognition characteristics (Cum100, CMCA, ROCA) of the weighted angle based distance were slightly improved.

**Table 4** Recognition accuracy using combined distance measure (number of features is 127)

| Method | Cum100, % | CMCA $[0..10^4]$ | First1, % | EER, % | ROCA $[0..10^4]$ |
|---|---|---|---|---|---|
| Simpl. Mahalanobis | 23.9 | 54.57 | 87.94 | 3.07 | 45.19 |
| Weighted angle | 26.0 | 57.15 | 88.42 | 3.31 | 44.08 |
| Combined: Simplified Mahalanobis (25%) + Weighted angle | 24.8 | 56.81 | 88.42 | 3.31 | 41.79 |

Further experiments were performed using FERET database containing photographies of 1196 persons. For recognition we used frontal faces from fa and fb sets (1196 fa, 1195 fb) with different facial expressions. We say that faces from fa are of known persons and use these images for training, faces from fb were used for testing. Images were normalized using manually selected eye centres. Then we performed masking using elliptical mask and histogram equalization for unmasked face part. For WPD+PCA recognition we used templates of the size 128x128 and decomposition to l=3 level (64 approximations and details images). For training we used all 1196 fa images. Experiments using different wavelet bases and distance measures showed that the best results are achieved using sym16 wavelets (symmlets with filter length 16) and weighted angle. Some results are presented in Table 5. For comparison we also present results using conventional PCA (for training were used all 1196 fa images).

**Table 5** Face recognition accuracy using FERET database (1196 fa, 1195 fb)

| Method | Dist. meas. | Feat. num. | Cum100, % | CMCA $[0..10^4]$ | First1, % | EER, % | ROCA $[0..10^4]$ |
|---|---|---|---|---|---|---|---|
| WPDP | WA | 1000 | 72.74 | 27.23 | 88.37 | 1.92 | 23.32 |
| PCA | WA | 1000 | 75.33 | 29.51 | 87.70 | 1.67 | 20.54 |
| WPDP | WHA | 300 | 59.45 | 34.18 | 87.11 | 2.59 | 30.87 |
| PCA | WHA | 300 | 81.61 | 48.31 | 88.62 | 2.09 | 39.05 |

Notation in table: "method" – used recognition method (WPDP - WPD+PCA with sym16 wavelets, PCA – traditional PCA); "dist. meas." – distance measure (WA - weighted angle, WHA – whitened angle); "feat. num." – number of features.

Also we performed experiments using WPD+PCA and combination of distance measures. At first we performed recognition using simplified Mahalanobis distance (numerator of weighted angle based distance), then resorted 1% of most similar images using weighted angle based distance. The results showed that using such combination we can perform recognition 1.67 times faster than using weighted angle based distance and achieve similar recognition accuracy. The results are presented in Table 6.

**Table 6** Recognition accuracy using WPD+PCA (sym16 wavelet, 1000 features) and combination of distance measures

| Distance measure | Cum100, % | CMCA $[0..10^4]$ | First1, % | EER, % | ROCA $[0..10^4]$ |
|---|---|---|---|---|---|
| Simpl. Mahalanobis | 76.76 | 30.70 | 84.10 | 2.51 | 37.76 |
| Weighted angle | 72.74 | 27.23 | 88.37 | 1.92 | 23.32 |
| Combined distance: simpl. Mahal. + weighted angle | 76.76 | 29.36 | 88.37 | 2.43 | 27.57 |

*The second subsection.* ("Face recognition results using automatically detected faces") Recognition experiments with automatically detected faces were performed using FERET database (1196 fa and 1195 fb images). For detection of faces and eyes we used normalized correlation based pre-detector and SNoW based verificator. For recognition we used pairs (fa, fb) of face images with correctly detected eyes. Recognition was performed using WPD+PCA, sym16 wavelets, 1000 features, and weighted angle based distance measure. Some results using different eye detection accuracy thresholds (and the number of image pairs) are presented in Table 7. Also we present recognition results with the same images using manually selected eye centres. The results showed that using automatically detected eye centres recognition accuracy is smaller than using manually selected eye centres.

**Table 7** WPD+PCA recognition accuracy using automatically and manually detected eyes

| Method | Det. thr., % | DB size | Cum100, % | CMCA $[0..10^4]$ | First1, % | EER, % | ROCA $[0..10^4]$ |
|---|---|---|---|---|---|---|---|
| auto. | 5 | 219 | 32.42 | 65.68 | 92.68 | 1.83 | 23.03 |
| manual | - | 219 | 6.85 | 51.81 | 95.89 | 1.37 | 7.09 |
| auto. | 6 | 386 | 41.19 | 82.05 | 86.79 | 3.89 | 64.45 |
| manual | - | 386 | 14.51 | 35.30 | 94.30 | 1.55 | 11.33 |
| auto. | 7 | 555 | 35.32 | 70.40 | 82.16 | 4.14 | 60.13 |
| manual | - | 555 | 12.43 | 27.21 | 92.43 | 1.44 | 11.13 |
| auto. | 8 | 686 | 38.48 | 63.60 | 81.05 | 3.94 | 57.19 |
| manual | - | 686 | 11.08 | 23.78 | 90.96 | 1.75 | 12.04 |
| auto. | 15 | 974 | 58.62 | 102.09 | 71.25 | 4.83 | 99.57 |
| manual | - | 974 | 17.45 | 21.98 | 88.71 | 1.85 | 15.37 |

Notation in table: "method" – used eye detection method ("auto." – detected automatically, "manual" – selected manually); "det. thr." – allowed threshold of the eye detection error; "DB size." - number of registered known persons in the database.

# Conclusions

1. Proposed modification of the fast normalized correlation -based face pre-detection method. The essence of the method is that image patterns are passed to the detector not one by one, as in traditional method, but all the patterns at once. The experiments show, that proposed modification speeds-up computation by 1.2-2.8 times, when using 5-20 image patterns.

2. Proposed a method for the detection of face contour in still greyscale images, based on active contour model. The essence of the method is that (1) contour is initialized inside the face using already detected facial features and then (2) initial contour is expanded using generalized gradient field. This allows to reduce influence of the scene background into face detection process and reduce contour detection errors.

3. Proposed new face recognition method, based on merging Wavelet Packet Decomposition and Principle Component Analysis methods. Using the proposed method, training time is almost independent from the number of training images, so this method could be used with large number of training images. Investigation showed, that training speed depends on (1) the size of training images and on (2) chosen number of wavelet packet decomposition levels.

4. Comparison of the proposed face recognition method with the traditional Principal Component Analysis -based method showed, that proposed method performs training faster and the differences of accuracy are small (<1%). The proposed method achieves highest accuracy when using for comparison of feature vectors weighted angle -based distance measure, traditional method - when using angle between "whitened" feature vectors -based distance measure.

5. Performed comparison of the proposed and known face recognition methods by accuracy, when using for recognition different distance measures between feature vectors. Comparison showed, that recognition accuracy is influenced not only by (1) extraction of features and by (2) selection of main features between extracted features, but also by (3) the distance measure, used for comparison of main features.

6. Proposed cumulative recognition characteristics - based method, that allows to evaluate if there is point for sequential merging of two chosen recognition algorithms. Algorithms are merged in order to increase recognition accuracy and speed. The essence of the proposed method is that accuracy and speed of the merged recognition algorithm are evaluated without long lasting experiments with the merged algorithm. The accuracy and speed of the merged algorithm are evaluated using the characteristics of its components (merged algorithms): accuracy and speed. The experiments showed, that the merging method allows to increase recognition accuracy and speed.

# The list of publications on the dissertation theme Year 2000-2004

Articles in editions included in the list of the Institute of Scientific Information (ISI):

1. Perlibakas, Vytautas. (2004). Distance measures for PCA-based face recognition. Pattern Recognition Letters, vol. 25, no. 6, 2004: pp. 711-724, ISSN 0167-8655.
2. Perlibakas, Vytautas. (2004). Face recognition using Principal Component Analysis and Wavelet Packet Decomposition. International Journal Informatica, vol. 15, no. 2, 2004: pp. 243-250, ISSN 0868-4952.
3. Perlibakas, Vytautas. (2003). Automatical detection of face features and exact face contour. Pattern Recognition Letters, vol. 24, no. 16, 2003: pp. 2977-2985, ISSN 0167-8655.

Articles published in the Lithuanian science editions included in the list certified by the Department of Science and Studies:

1. Perlibakas, Vytautas. (2003). Distance measures for PCA-based face recognition. Periodical Journal Information Technology and Control, Lithuania, Kaunas, Technologija, vol. 29, no. 4, 2003: pp. 67-74, ISSN 12392-1215.
2. Perlibakas, Vytautas. (2003). Automatical Detection of Face Features and Exact Face Contour. Periodical Journal Information Technology and Control, Lithuania, Kaunas, Technologija, vol. 26, no. 1, 2003: pp. 67-71, ISSN 12392-1215.
3. Perlibakas, Vytautas. (2002). Image and geometry based face recognition. Periodical Journal Information Technology and Control, Lithuania, Kaunas, Technologija, vol. 22, no. 1, 2002: pp. 73-79, ISSN 12392-1215.

Publications in conference proceedings:

1. Perlibakas, Vytautas. (2004). Veido atpažinimas naudojant Karuno-Loevo, kosinusinę ir bangelių transformaciją. Konferencijos "Informacinės technologijos 2004" pranešimų medžiaga, Kaunas, 2004: pp. 117-128, ISBN 9955-09-588-1.
2. Puniene, Jurate, Punys, Jonas, Punys, Vytenis, Perlibakas, Vytautas, Vaitkevicius, Vytautas. (2002). Image processing and compression techniques with applications in medicine, surveillance and industry. In Proceedings of the International Conference "East - West Vision 2002" (EWV'02), International Workshop and Project Festival, Computer Vision, Computer Graphics, New Media. Franz Leberl, Andrej Ferko (eds.). Graz, September 12-13, 2002: pp. 215-216, ISBN 3-85403-163-7.
3. Perlibakas, Vytautas. (2002). Kontūrais ir elipsėmis pagrįstas veidų detektavimas. Tarptautinės konferencijos "Informatika sveikatos apsaugai '2002" pranešimų medžiaga. Visaginas, 2002: pp. 62-67, ISBN 9955-09-260-2.

Other publications:

Information about the author of the dissertation

Vytautas Perlibakas was born on May 20, 1976 in Vilkaviskis. In 1994 graduated from S. Neris secondary school at Vilkaviskis. During 1994-1998 studied at Kaunas University of Technology, Faculty of Informatics and obtained Bachelor's in Informatics qualifying degree. During 1998-2000 studied at Kaunas University of Technology, Faculty of Informatics and obtained Master's of Science in Informatics qualifying degree (diploma First Class). During 2000-2004 studied at Kaunas University of Technology and prepared doctoral dissertation "Computerized face detection and recognition".

# KOMPIUTERIZUOTAS VEIDO DETEKTAVIMAS IR ATPAŽINIMAS

Reziumė

Disertacijoje nagrinėjami kompiuterizuoti veido detektavimo, analizės ir atpažinimo metodai, siekiant padidinti jų tikslumą ir greitį, o sukurtus metodus ateityje panaudoti automatinėse veido atpažinimo sistemose. Disertaciją sudaro įvadas, 6 skyriai, išvados, bibliografinės nuorodos (306 nuorodos), autoriaus publikacijų disertacijos tema sąrašas. Disertacijos apimtis 186 puslapiai, 23 lentelės, 26 paveikslėliai.

Įvade aptariami darbo tikslai ir uždaviniai, tyrimui naudojami metodai, darbo mokslinis naujumas ir praktinė reikšmė. Kompiuterizuotas veido detektavimas ir atpažinimas yra kompiuterinės regos problemos, intensyviai tyrinėjamos jau daugiau kaip dešimt metų. Veido atpažinimas taip pat yra viena iš biometrinių technologijų (tokių kaip balso, pirštų antspaudų, akies rainelės atpažinimas), kurios gali būti naudojamos žmogaus tapatybės nustatymui. Biometrinių technologijų, tame tarpe ir veido atpažinimo, tyrimai yra atliekami siekiant sukurti patikimesnes duomenų apsaugos priemones bei ieškant naujų nusikalstamumo ir terorizmo prevencijos būdų. Veido atpažinimo technologija leidžia atlikti pasyvų asmens identifikavimą sudėtingoje aplinkoje. Todėl ji gali būti naudojama valstybės sienų bei kitų teritorijų apsaugai. Veido detektavimo, analizės ir atpažinimo metodai taip pat gali būti naudojami kriminalistikoje, kino pramonėje, medicinoje, kuriant intelektualią namų aplinką ir daugelyje kitų sričių. Potencialių taikymų gausa skatina šių metodų tyrimus. Kadangi veido detektavimo ir atpažinimo tikslumą įtakoja daug faktorių (keičiasi pats veidas ir jį supanti aplinka), problemos yra sudėtingos ir nėra pilnai išspręstos. Šiuolaikinė kompiuterinė technika leidžia sukaupti bazėse didelio skaičiaus asmenų veido vaizdus. Todėl reikalingi metodai, kurie leistų atlikti paiešką šiose bazėse. Taip pat reikia kurti greitus detektavimo ir atpažinimo metodus, kad jie atitiktų esamą kompiuterių spartą. Disertacijos tikslai: a) sukurti naujus arba patobulinti esamus veido detektavimo, analizės ir atpažinimo metodus, siekiant padidinti šių metodų greitį ir tikslumą; b) tirti veido požymių ir veido kontūro detektavimo problemas. Uždaviniai: 1. Ištirti esamus ir pasiūlyti naujus arba modifikuotus veido aptikimo vaizde ir veido požymių detektavimo bei veido atpažinimo metodus, kurie leistų pasiekti didesnį greitį ir tikslumą nei jau esami metodai. 2. Atlikti tokius tyrimus: a) ištirti ir įvertinti atstumo tarp požymių mato pasirinkimo įtaką atpažinimo tikslumui, naudojant diskrečiosiomis transformacijomis pagrįstus atpažinimo metodus; b) ištirti atpažinimo algoritmų apjungimo metodus, leidžiančius padidinti atpažinimo greitį ir tikslumą; c) ištirti ir modifikuoti greitam pradiniam veido aptikimui naudojamus metodus, pagrįstus kontūrais, matematine morfologija bei vaizdo pavyzdžiais; d) ištirti ir modifikuoti veido požymių detektavimo metodus, pagrįstus matematine morfologija, vaizdo profilių analize bei vaizdo pavyzdžiais; e) ištirti ir modifikuoti tikslaus veido kontūro detektavimo metodus, pagrįstus aktyviaisiais kontūrais. 3. Tyrimų rezultate suformuluoti rekomendacijas veido aptikimo ir atpažinimo kompiuterizuotų metodų plėtrai.

Pirmame disertacijos skyriuje apžvelgiama literatūra veido ir veido požymių detektavimo bei veido atpažinimo tema. Pateikiama metodų klasifikacija, aptariamos metodų palyginimui naudojamos charakteristikos, pateikiama informacija apie veidų bazes, kurios naudojamos detektavimo ir atpažinimo metodų apmokymui bei testavimui. Taip apžvelgiama, kokiose srityse gali būti naudojami veido analizės ir atpažinimo metodai, jų naudojimo scenarijai. Šiame skyriuje taip pat aptariami stambesni veido atpažinimo sistemų kūrimo projektai, jų komercializavimo istorija ir ryšys su universitetais ir kitomis mokslo institucijomis.

Antrame skyriuje sudaromi ir tiriami pradinio veido detektavimo metodai: sukurtas kontūrais ir elipsėmis pagrįstas metodas ir siūlomas greita normalizuota koreliacija tarp kelių vaizdo pavyzdžių pagrįstas metodas. Analizuojami pradinių detektavimų verifikavimui naudojami vaizdu pagrįsti metodai bei euristikos, naudojamos persidengiančių detektavimų atmetimui.

Trečiame skyriuje analizuojamas matematine morfologija pagrįstas veido požymių detektavimo metodas bei projekcijų analize ir vaizdo pavyzdžiais pagrįstas akių detektavimo metodas.

Ketvirtame skyriuje siūlomas aktyviais kontūrais, veido požymiais ir apibendrintu gradientiniu lauku pagrįstas tikslaus veido kontūro detektavimo metodas.

Penktame skyriuje aptariami diskrečiosiomis transformacijomis (Karuno-Loevo, kosinusine, bangelių) pagrįsti veido atpažinimo metodai. Analizuojami svarbiausių požymių atrinkimo bei jų palyginimo metodai. Šiame skyriuje siūlomas naujas atpažinimo metodas, pagrįstas bangelių paketų dekompozicija ir pagrindinių dedamųjų analize. Taip pat siūlomas kumuliatyvinėmis atpažinimo charakteristikomis pagrįstas atpažinimo algoritmų apjungimo metodas.

Šeštame skyriuje analizuojami eksperimentų metu gauti veido detektavimo, atpažinimo ir veido kontūrų detektavimo rezultatai. Pateikiami atpažinimo rezultatai naudojant įvairius atpažinimo metodus, rankiniu ir automatiniu būdu detektuotus požymius. Gauti rezultatai palyginami tarpusavyje ir su kitų autorių rezultatais.

Atlikus disertacijoje aprašytus tyrimus ir eksperimentus, padarytos tokios išvados:

1. Pasiūlyta greitos normalizuotos koreliacijos metodo modifikacija pradiniam veido aptikimui. Modifikacijos esmė yra ta, kad vaizdo pavyzdžiai yra paduodami grupėmis, o ne po vieną, kaip įprastiniame metode. Eksperimentai rodo, kad modifikacija pagreitina skaičiavimus 1.2-2.8 karto, kai naudojama 5-20 vaizdo pavyzdžių.

2. Pasiūlytas naujas veido kontūro detektavimo pilkaspalviuose vaizduose metodas, kuris remiasi aktyviaisiais kontūrais. Metodo esmė - (1) inicializuoti pradinį kontūrą veido viduje panaudojant jau detektuotus veido požymius ir (2) išplėsti pradinį kontūrą panaudojant apibendrintą gradientinį lauką. Pasiūlytas metodas leidžia sumažinti vaizdo fono įtaką veido kontūro detektavimui ir, kaip rodo eksperimentai, leidžia sumažinti kontūro detektavimo klaidas.

3. Pasiūlytas naujas veido atpažinimo metodas, pagrįstas bangelių paketų dekompozicijos ir pagrindinių dedamųjų analizės metodų apjungimu. Kadangi naudojant šį metodą apmokymo laikas beveik nepriklauso nuo vaizdų imties, pasiūlytas metodas taikytinas esant dideliam apmokymo vaizdų skaičiui. Nustatyta, kad metodo apmokymo greitis priklauso nuo (1) vaizdų dydžio ir (2) nuo pasirinkto bangelių paketų dekompozicijos lygmenų skaičiaus.

4. Pasiūlyto veido atpažinimo metodo palyginimas su įprastiniu pagrindinių dedamųjų analizės metodu parodė, kad pasiūlytu metodu apmokymas atliekamas greičiau, o metodų tikslumas skiriasi nežymiai (<1%). Pasiūlytas metodas didžiausią atpažinimo tikslumą pasiekia požymių palyginimui naudojant kampu su svoriais pagrįstą atstumo matą, įprastinis metodas - naudojant kampu tarp "baltintų" požymių pagrįstą matą.

5. Atliktas pasiūlyto ir žinomų atpažinimo metodų palyginimas pagal tikslumą, kai atpažinimui naudojami skirtingi atstumo tarp požymių matai. Palyginimas parodė, kad atpažinimo tikslumui įtaką daro ne tik tai, (1) kokie požymiai išskiriami ir (2) kaip atrenkami svarbiausi iš jų, bet ir tai, (3) kokį atstumo matą naudojant atrinktųjų požymių vektoriai yra palyginami.

6. Pasiūlytas kumuliatyvinėmis atpažinimo charakteristikomis pagrįstas metodas, kuris leidžia įvertinti ar tikslinga nuosekliai apjungti du pasirinktus atpažinimo algoritmus. Algoritmų apjungimo tikslas yra pasiekti didesnį atpažinimo tikslumą ir greitį. Pasiūlyto metodo esmė yra ta, kad apjungtojo algoritmo tikslumas ir greitis yra įvertinami neatliekant ilgai trunkančių eksperimentų su naujai sukonstruotu jungtiniu algoritmu. Naujojo jungtinio algoritmo tikslumas ir greitis yra įvertinami pagal jo sudedamųjų dalių žinomas charakteristikas: tikslumą ir greitį. Eksperimentiškai patikrinus teorinius įvertinimus yra įsitikinta, kad apjungimo metodas leidžia padidinti tiek atpažinimo tikslumą, tiek ir greitį.

Informacija apie disertacijos autorių

Vytautas Perlibakas gimė 1976 m. gegužės 20 d. Vilkaviškyje. 1994 m. baigė Vilkaviškio S. Nėries vidurinę mokyklą. 1994-1998 m. studijavo Kauno technologijos universitete, Informatikos fakultete ir įgijo informatikos bakalauro kvalifikacinį laipsnį. 1998-2000 m. studijavo Kauno technologijos universitete, Informatikos fakultete ir įgijo informatikos magistro kvalifikacinį laipsnį (diplomas su pagyrimu). 2000-2004 m. Kauno technologijos universitete, studijuodamas fizinių mokslų srities informatikos krypties doktorantūroje, parengė daktaro disertaciją tema "Kompiuterizuotas veido detektavimas ir atpažinimas".